\documentclass[5p]{elsarticle}
\usepackage{booktabs,shortvrb}
\usepackage{amsmath}
\usepackage{amsfonts}%
\usepackage{amssymb}%
\usepackage{booktabs}
\usepackage{url}
\usepackage{algpseudocode}
\usepackage{algorithm}

\makeatletter
    \def\ps@pprintTitle{%
      \let\@oddhead\@empty
      \let\@evenhead\@empty
      \let\@oddfoot\@empty
      \let\@evenfoot\@oddfoot
    }
\makeatother
\begin{document}

\begin{frontmatter}
\title{Sensitive Ants in Solving the Generalized Vehicle Routing Problem}
\author[1]{Camelia-M. Pintea}\ead{cmpintea@cs.ubbcluj.ro}
\author[1]{Camelia Chira}\ead{chira@cs.ubbcluj.ro}
\author[1]{D. Dumitrescu}\ead{ddumitr@cs.ubbcluj.ro}
\author[2]{Petrica C. Pop}\ead{pop\_petrica@yahoo.com }
\address[1]{Department of Computer Science, Babes-Bolyai University, Cluj-Napoca 400084, Romania}
\address[2]{North University of Baia Mare, Romania, V. Babe\c s, 430083, Baia Mare}

\begin{abstract}
The idea of sensitivity in ant colony systems has been exploited in hybrid ant-based models with promising results for many combinatorial optimization problems. Heterogeneity is induced in the ant population by endowing individual ants with a certain level of sensitivity to the pheromone trail. The variable pheromone sensitivity within the same population of ants can potentially intensify the search while in the same time inducing diversity for the exploration of the environment. The performance of sensitive ant models is investigated for solving the generalized vehicle routing problem. Numerical results and comparisons are discussed and analysed with a focus on emphasizing any particular aspects and potential benefits related to hybrid ant-based models. 
\end{abstract}

\end{frontmatter}

%%%%%%%%%%%%%%%%%%%%%%%%%
\section{Introduction}
The potential of ant-based models \cite{Dorigo2005,Dorigo1997,Pintea2008} in solving difficult optimization problems has been well emphasized by successful results obtained in many and varied fields including transportation optimization, quadratic assignment, scheduling, vehicle routing and protein folding. Inspired by the real-world collective behaviour of social insects, {\it Ant Colony System (ACS)} algorithms \cite{Dorigo2005} rely on the stigmergic interactions between many identical artificial ants to find solutions to a given problem. Each ant generates a complete tour (associated to a problem solution) by probabilistically choosing the next node at each path intersection based on the cost and the amount of pheromone on the connecting edge. Stronger pheromone trails are preferred and the most promising tours build up higher amounts of pheromone in time.

Inducing heterogeneity in the population by enabling each artificial ant to react in a different way to the same environment \cite{Pintea2009} represents a promising approach to the application of ant-based models for solving complex real-world problems possibly with a dynamic character. Each individual ant can be endowed with a certain level of sensitivity to the pheromone trail triggering various types of reactions to a changing environment. The variable pheromone sensitivity within the same population of ants can potentially intensify the search (normally through high sensitivity levels) while in the same time inducing diversity for the exploration of the environment. The decision of a low-level sensitive ant regarding the action to be performed crucially contributes to the quality of the search process and solutions. 

The Generalized Vehicle Routing Problem (GVRP) is an extension of the Vehicle Routing Problem (VRP) and was introduced by Ghiani and Improta \cite{Ghiani2000}. The GVRP is the problem of designing optimal delivery or collection routes, subject to capacity restrictions, from a given depot to a number of predefined, mutually exclusive and exhaustive node-sets (clusters). 

The GVRP belongs to the class of generalized combinatorial optimization problems, which are natural extensions of combinatorial optimization problems by considering a related problem relative to a given partition of the nodes of the graph into node sets, while the feasibility constraints are expressed in terms of the clusters. In the literature we can find several generalized problems such as the generalized minimum spanning tree problem (see \cite{Pop2009c}), the generalized traveling salesman problem, the generalized vehicle routing problem, the generalized (subset) assignment problem, etc. These generalized problems belong to the class of NP-complete problems, are harder than the classical ones and nowadays are intensively studied due to the interesting properties and applications in the real world, even though many practitioners are reluctant to use them for practical modeling problems because of the complexity of finding optimal or near-optimal solutions.

Ghiani and Improta \cite{Ghiani2000} showed that the problem can be transformed into a capacitated arc routing problem (CARP) and Baldacci et al. \cite{Baldacci2008} proved that the reverse transformation is valid. Recently, Pop \cite{Pop2009b} provided a new efficient transformation of the GVRP into the classical vehicle routing problem (VRP). As far as we know, the only specific algorithm for solving the GVRP was developed by Pop et al. \cite{Pop2009a} and was based on ant colony optimization.

The aim of this paper is to investigate the performance of the {\em Sensitive Ant Model (SAM)} \cite{Pintea2009}  in solving the {\it Generalized Vehicle Routing Problem (GVRP)} \cite{Ghiani2000}. We report numerical results of the {\it SAM} model for several {\it GVRP} benchmark problems and discuss the performance of {\it SAM} compared to the standard {\it ACS} technique.

\section{Definition and Complexity of the GVRP}

Let $G = (V,A)$ be a directed graph with $V = \{0, 1, 2, .... ,n\}$ as the set of vertices and the set of arcs $A=\{ (i,j)\;|\;i,j \in V, i\neq j \}$. A nonnegative cost $c_{ij}$ associated with each arc $(i,j)\in A$. The set of vertices (nodes) is partitioned into $k+1$ mutually exclusive nonempty subsets, called clusters, $V_0, V_1,..., V_k$ (i.e. $V = V_0 \cup  V_1 \cup ... \cup  V_k$ and $V_l \cap  V_p = \emptyset$  for all $l,p\in \{0,1,...,k\}$ and $l\neq p$). The cluster $V_0$ has only one vertex $0$, which represents the depot, and remaining n nodes belonging to the remaining $k$ clusters represent geographically dispersed customers. Each customer has a certain amount of demand and the total demand of each cluster can be satisfied via any of its nodes. There exist $m$ identical vehicles, each with a capacity $Q$.

\medskip
 
\begin{figure}[tbhp]
\centering
\includegraphics[scale=0.40]{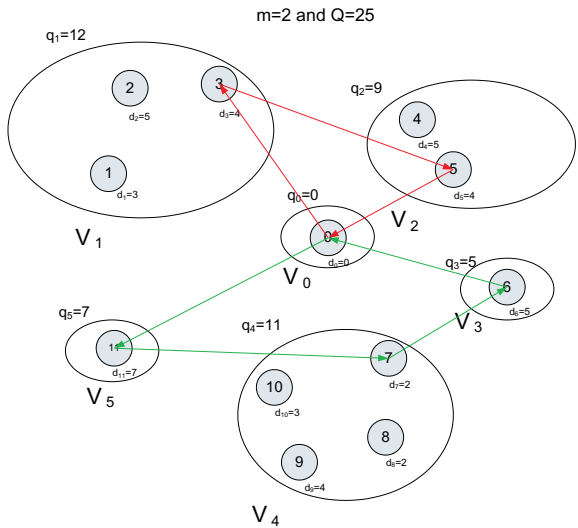}
\caption{An example of a feasible solution of the GVRP}
\label{gvrp}
\end{figure}

The generalized vehicle routing problem (GVRP) consists in finding the minimum total cost tours of starting and ending at the depot, such that each cluster should be visited exactly once, the entering and leaving nodes of each cluster is the same and the sum of all the demands of any tour (route) does not exceed the capacity of the vehicle $Q$. An illustrative scheme of the GVRP and a feasible tour is shown in the Figure \ref{gvrp}.

The GVRP reduces to the classical Vehicle Routing Problem (VRP) when all the clusters are singletons and to the Generalized Traveling Salesman Problem (GTSP) when $m = 1$ and $Q =\infty$.

The GVRP is $NP$-hard because it includes the generalized traveling salesman problem as a special case when $m=1$ and $Q=\infty$.

\medskip

Several real-world situations can be modelled as a GVRP. The post-box collection problem described in Laporte et al. \cite{Laporte1989} becomes an asymmetric GVRP if more than one vehicle is required. Furthermore, the GVRP is able to model the distribution of goods by sea to a number of customers situated in an archipelago as in Philippines, New Zeeland, Indonesia, Italy, Greece and Croatia. In this application, a number of potential harbours is selected for every island and a fleet of ships is required to visit exactly one harbour for every island.

\medskip

Several applications of the GTSP (Laporte et al. \cite{Laporte1996}) may be extended naturally to GVRP. In addition, several other situations can be modeled as a GVRP, these include:

\begin{itemize}

\item the Traveling Salesman Problem (TSP) with profits (Feillet et al. \cite{Feillet2005});

\item a number of Vehicle Routing Problem (VRP) extensions: the VRP with selective back hauls, the covering VRP, the periodic VRP, the capacitated general windy routing problem, etc.; 

\item the design of tandem configurations for automated guided vehicles (Baldacci et al. \cite{Baldacci2008}).

\end{itemize}

\section{The ACS algorithm for solving GVRP}
The ACS-based algorithm for GVRP  \cite{Pop2009a} uses artificial ants in order to construct vehicle routes by successively choosing exactly one node from each cluster. This task continues until each cluster has been visited. Whenever the choice of another node from a cluster would lead to an infeasible solution because of vehicles capacity, the depot is chosen and a new route is started.

Initially, the Nearest Neighbor (NN) algorithm - with the rule always go to the nearest as-yet-unvisited location - is considered. The best solution of Nearest Neighbor (NN) algorithm ($L^+$) is used for ACS-based algorithm start.

The number of ants corresponds to the number of GVRP customers {\it m}. At the beginning of an iteration, an ant is placed at each node (customer). After initializing the basic ant system algorithm, the two steps: (i) construction of vehicle routes and (ii) trail update are repeated for a given number of iterations.

To favor the selection of an edge with a high pheromone level and high visibility, a probability function $p_{k}^{ij}$ is defined as follows:

\begin{equation}
	p_{k}^{ij}(t) =\frac{\tau_{ij}^k(t)[\eta_{ij}^k(t)]^\beta}{\sum_{o\in J^k_{i}}\tau_{io}^k(t)[\eta_{io}^k(t)]^\beta}
	\label{eq_acs}	
\end{equation}
where $J_i^k$ is the set of unvisited neighbors of node $i$ by ant $k$, $j \in J_i^k$ and $\beta$ is a parameter used for tuning the relative importance of visibility. 

After an artificial ant has constructed a feasible solution, the pheromone trails are laid depending on the objective value $L_k$. For each edge that was used by ant $k$, the pheromone trail is updated according to the following rule:
\begin{equation}
\tau_{ij}(t+1)=(1-\rho)\tau_{ij}(t)+\rho\frac{1}{L^{k}}
\end{equation}
where $\rho \in (0, 1)$ is an evaporation rate parameter.

A tabu list prevents ants visiting clusters they have previously visited. The ant tabu list is cleared after each completed tour.

The global update rule, applied by the elitist ants, as in ACS \cite{Dorigo1997} is:
\begin{equation}
\tau_{ij}(t + 1) = (1 - \rho)\tau_{ij}(t) + \rho\frac{1}{L^{+}},
\end{equation}
where $L^+$ is the so far best solution.

\section{Sensitive Ant-based Model for GVRP}

The {\it Sensitive Ant Model (SAM)} technique proposed in \cite{Pintea2009} is engaged for solving the {\it GVRP}. The general approach to solving GVRP using SAM is the same with the ACS approach presented in the previous section except that the transition probabilities defined by SAM are used. The initialization of the algorithm, the update rules and the maintenance of a tabu list are kept the same in SAM for GVRP.

The {\it SAM} algorithm involves several ants able to communicate in a stigmergic manner (influenced by pheromone trails) for solving complex search problems. Within the {\it SAM} model, each ant is characterized by a pheromone sensitivity level ({\em PSL}). The {\em PSL} value is expressed by a real number in the unit interval [0, 1]. When {\em PSL} is null the ant completely ignores stigmergic information and when {\em PSL} is one the agent has maximum pheromone sensitivity. The ants with a low {\em PSL} value are more independent and are considered environment explorers. They have the potential to autonomously discover new promising regions of the solution space. The ants with high {\em PSL} values are very sensitive to pheromone traces. They are influenced by stigmergic information and therefore intensively exploit the promising search regions already identified.

SAM introduces a measure of randomness proportional to the level of individual {\em PSL} in the decisions of ants regarding the path to follow. This is achieved by modifying the transition probabilities using the {\em PSL} values in a renormalization process \cite{Pintea2009}.  The SAM renormalized transition probability for ant $k$ (influenced by {\em PSL}) is denoted by $sp_{k}^{ij}(t)$ and is given by the following equation:

\begin{equation}
sp_{k}^{ij}(t) = p_{k}^{ij}(t) \cdot PSL_{k}(t) ,
\label{eq:nou}
\end{equation}

where $p_{k}^{ij}(t)$ is the probability for ant {\em k} to choose the next node {\em j} from current node {\em i} (as given in {\em ACS} - see Equation \ref{eq_acs}) and $PSL_{k}(t)$ represents the {\em PSL} value of ant {\em k} at time {\em t}.

It can be noticed that the SAM probability of selecting the next node is the same with the ACS one when PSL value is one. In order to associate a standard probability distribution to the system, the SAM virtual state corresponding to the 'lost' probability of $(1-PSL_{k}(t))$ has to be defined. The associated virtual state decision rule specifies the action to be taken when the virtual state is selected using the renormalized transition mechanism. The following rule is used in the current paper: the ant randomly chooses an available node with uniform probability if the virtual state is selected. This approach favors the increasing of randomness in the selection process with the decreasing of sensitivity level to pheromone. 

\section{Computational results}

The performance of the SAM and ACS for solving {\it GVRP} is investigated. Numerical experiments focus on seven benchmark problems from the {\it TSPLIB} library \cite{vrplib}. These problems contain between 51 and 101 customers (nodes), which are partitioned into a given number of clusters, and in addition the depot.

Originally the set of nodes in these problems is not divided into clusters. The CLUSTERING procedure proposed by Fischetti {\it et al.} \cite{Fischetti1997} is used to divide data into node-sets. This procedure sets the number of clusters $m=[\frac{n}{5}]$, identifies the $m$ farthest nodes from each other and assigns each remaining node to its nearest center.

Table~\ref{tab:problems} contains the description of the GVRP instances addressed in this paper. 
\begin{table}
\begin{center}
\caption{Problem characteristics for the ant-based algorithms for {\it GVRP}}
\vspace{2pt}
\label{tab:problems}
\begin{tabular}{|c|c|c|c|c|c|}
\hline {\it Problem} & {\it VR} & {\it Q}& {\it Q'} & {\it No.vehicles} & {\it No.Routes} \\
\hline\hline 
11eil51 & 2 &160&320&  6 & 3  \\
\hline
16eil76A &  2 & 140&280 & 10 & 5   \\
\hline
16eil76B & 3& 100&300  & 15 & 5   \\
\hline
16eil76C &  2& 180&360& 8 & 4  \\
\hline
16eil76D &  2&220&440& 6 & 3 \\
\hline
21eil101A &2&200&400&  8 & 4   \\
\hline
21eil101B &  2& 112&224& 14 & 7   \\
\hline
\end{tabular}
\end{center}
\end{table}

The meaning associated with the columns in Table~\ref{tab:problems} is as follows:

\begin{itemize}
\item Problem: The name of the test problem contains the number of clusters (first digits in the problem name) and the number of nodes (last digits in the problem name).
\item VR: The minimal number of vehicles needed for a route in order to cover even the largest capacity of a cluster (VR=Vehicles/Route)
\item	Q': the capacity $Q\cdot VR$, where $Q$ is the capacity of a vehicle available at the depot.
\end{itemize}

The same parameter setting was used in both SAM and ACS algorithms in order to allow a meaningful direct comparison: $\tau_{0}=0.1$ (the initial value of all pheromone trails), $\alpha=1$, $\beta=5$, $\rho=0.0001$ and $q_{0}=0.5$. In the SAM algorithm, the $PSL$ value is randomly generated between 0 and 1 for each ant.

\begin{figure}[tbhp]
\centering
\includegraphics[scale=0.35]{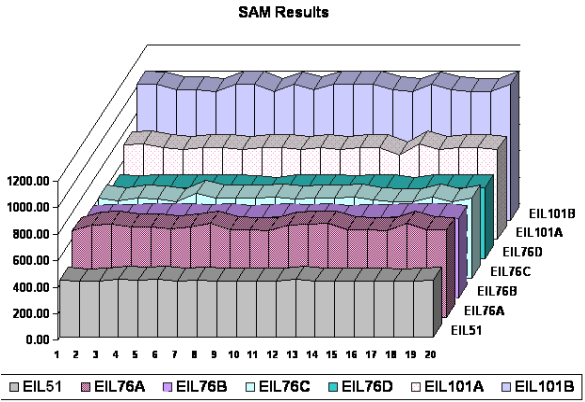}
\caption{SAM results from 20 runs for the considered problem instances \cite{vrplib}}
\label{sbfig}
\end{figure}

Numerical results indicate a competitive performance of the SAM algorithm. Figure \ref{sbfig} presents SAM results from 20 successive runs for the considered problem instances.

Tables~\ref{tab:best-results} and~\ref{tab:avg-results} present comparative numerical results obtained in $20$ runs. The performance of SAM in solving GVRP is compared to that of ACS \cite{Pop2009a}. 

The following information is contained in Tables~\ref{tab:best-results} and~\ref{tab:avg-results}:
\begin{itemize}
\item Best length: the minimal length of collection routes;
\item Best time: the time of the minimal collection routes;
\item Avg. length: the average length for 20 runs;
\item Avg. time: the average time (in seconds) for 20 runs.
\end{itemize}

\begin{table}
\begin{center}
\caption{Best Values and Times - {\it ACS} and {\it SAM} algorithms for solving {\it GVRP}}
\vspace{2pt}
\label{tab:best-results}
\begin{tabular}{|c|c|c|c|c|}
\hline {\it Problem} &{\it ACS}&{\it Time ACS }&{\it SAM}&{\it Time SAM }\\
\hline\hline 
11eil51 &  418.85& {\bf 212}&{\bf 418.21}& 297  \\
\hline
16eil76A & 668.78 & {\bf 18} &  {\bf 651.98}& 25 \\
\hline
16eil76B & 625.83 & {\bf 64 } & {\bf 599.23 }& 166 \\
\hline
16eil76C  & {\bf 553.21}  & 215.00 &  577.49& {\bf 88}\\
\hline
16eil76D & {\bf 508.81} & 177.00& 515.64 & {\bf 120 }\\
\hline
21eil101A & {\bf 634.74} &{\bf 72} & {\bf 634.74} &  111\\
\hline
21eil101B & {\bf 875.58} & {\bf 8.00} & 966.17 & 52  \\
\hline
\end{tabular}
\end{center}

\end{table}

\begin{table}
\begin{center}
\caption{Average Values and Times - {\it ACS} and {\it SAM} algorithms for solving {\it GVRP}}
\vspace{2pt}
\label{tab:avg-results}
\begin{tabular}{|c|c|c|c|c|}
\hline {\it Problem} &{\it ACS}&{\it Time ACS }&{\it SAM}&{\it Time SAM }\\
\hline\hline 
11eil51 & 429.85 &  210.20 & {\bf 424.05 }& {\bf  96.40} \\
\hline
16eil76A & 706.09 & {\bf 109.20}& {\bf 677.85} &  187.60\\
\hline
16eil76B & 684.04 & {\bf 50.7} & {\bf 608.62} & 173.7 \\
\hline
16eil76C  & 625.87  & 73.00 & {\bf 602.06} & {\bf 42.25}\\
\hline
16eil76D & 566.56 &{\bf 93.20} & {\bf 533.12} & 223.45 \\
\hline
21eil101A & 699.46 & {\bf 29.00}& {\bf 690.39} & 124.30 \\
\hline
21eil101B &{\bf 996.41} & {\bf 25.95} &  998.71& 27.90  \\
\hline
\end{tabular}
\end{center}

\end{table}

The computational values are the result of the average of $20$ successively runs of both algorithms. Termination criterion is either the maximum number of iterations, $N_{iter}=250000$ or the maximum running time (five minutes) on a AMD 2600, 1.9Ghz and 1024 MB. 

When comparing the best solution reported in $20$ runs, the performance of the ACS and SAM algorithms are similar. SAM reports better values for three out seven problems while ACS does better for other three problems. It should be noticed that whenever a method is able to obtain a better solution, it also reports a longer running time compared to the other one.

The average solutions obtained by SAM are clearly improved compared to ACS although the running time has slightly increased for some of the GVRP instances. SAM detects a better average solution (calculated based on the 25 runs) for six out of the seven benchmark problems. Overall better average SAM results are facilitated by a better exploration of the search space and exploitation of new solutions. This is due to the variable sensitivity induced in SAM via random individual PSL values. 

\section{Conclusions}
Sensitive heterogeneous ant-based models facilitate a balanced search process by endowing ants with different pheromone sensitivity levels translated into different search strategies. An effective exploration of the search space is performed particularly by ants having low pheromone sensitivity while the exploitation of intermediary solutions is facilitated by highly-sensitive ants.

The performance of hybrid ant-based models is investigated with successful results for solving the NP-hard Generalized Vehicle Routing Problem. Variable pheromone sensitivity in ant-based models proves to be benefic to the search process leading to better results compared to the ant colony system algorithm. Numerical results encourage the exploration of new ways to induce heterogeneity in ant-based models. \\

\bigskip

\section*{Acknowledgements.}
This work was supported by CNCSIS UEFISCSU, project number PNII  IDEI 508/2007 \textit{New Computational Paradigms for Dynamic Complex Problems}.

\end{document}